\documentclass{article}
\usepackage{ijcai23}

\usepackage{xcolor}

\usepackage{times}
\usepackage{soul}
\usepackage{url}
\usepackage[hidelinks]{hyperref}
\usepackage[utf8]{inputenc}
\usepackage[small]{caption}
\usepackage{graphicx}
\usepackage{amsmath}
\usepackage{amsthm}
\usepackage{booktabs}
\usepackage{algorithm}
\usepackage{algorithmic}
\usepackage[switch]{lineno}

\title{An All Deep System for Badminton Game Analysis \\

{\footnotesize Golden Award for IJCAI CoachAI Challenge 2023: Team NTNUEE AIoTLab}
}

\author{
Po-Yung Chou
\and
Yu-Chun Lo\and
Bo-Zheng Xie\and
Cheng-Hung Lin$^*$\And
Yu-Yung Kao
\affiliations
National Taiwan Normal Unversity\\ % $^1$
\emails
corresponding author: $^*$brucelin@ntnu.edu.tw
}
\begin{document}

\maketitle

\begin{abstract}
    The CoachAI Badminton 2023 Track1 initiative aim to automatically detect events within badminton match videos. Detecting small objects, especially the shuttlecock, is of quite importance and demands high precision within the challenge. Such detection is crucial for tasks like hit count, hitting time, and hitting location. However, even after revising the well-regarded shuttlecock detecting model, TrackNet, our object detection models still fall short of the desired accuracy. To address this issue, we've implemented various deep learning methods to tackle the problems arising from noisy detectied data, leveraging diverse data types to improve precision. In this report, we detail the detection model modifications we've made and our approach to the 11 tasks. Notably, our system garnered a score of 0.78 out of 1.0 in the challenge. We have released our source code in Github \url{https://github.com/jean50621/Badminton_Challenge}
    
\end{abstract}

\section{Introduction}

The objectives of CoachAI Badminton 2023 Track1\cite{task1} are described as follows. Participants are required to construct a computer vision system that generates 11 targets from an input video. For clarity, the 11 tasks are listed below:

\begin{description}
   \item[Number of shots] The total number of shots within a match, which is equivalent to the number of hits. It's crucial to note that if the predicted number of shots deviates from the ground truth, the task will be scored as 0 points.
    
    \item[Hit frame] This denotes the precise moment of hitting in the video. Any prediction error exceeding 2 frames will be deemed incorrect.
    
    \item[Hitter] The task is to identify which player strikes the shuttlecock.
    
    \item[BallHeight] This refers to the altitude of the hitting point.
    
    \item[RoundHead] Classify whether the player's action was a round head shot or not during the hit.
    
    \item[Backhand] Determine if the player used a backhand action during the hit.
    
        \item[LandingX/LandingY] This represents the shuttlecock's projected position on the court at the end of its trajectory, whether struck by the opponent or when it touches the ground, measured in pixels.
    
    \item[HitterLocationX/HitterLocationY] This signifies the position of the player executing the strike, using the toe of their lead foot as the coordinate reference in the image, in pixels. If the player jumps, the projected coordinates onto the court are considered.
    
    \item[DefenderLocationX/DefenderLocationY] This denotes the position of the player on the defensive, and is determined in a manner similar to the hitting player's location.

    \item[BallType] There are nine types of balls, labeled from 1 to 9. For additional details, refer to: \url{https://sites.google.com/view/coachai-challenge-2023/tasks/track1}
    
    \item[Winner] This task involves identifying the winning player. 
    
\end{description}

To construct a system capable of addressing all these tasks, we've adopted a two-step process. The first and foundational function for these tasks is the object detection model, which is associated with aspects like location projection, the trajectory of the object, and the object's precise location, among others. After determining the object's location, a post-processing model is employed to produce the predictions for different task. In the subsequent two sections, we will delve into the object detection techniques in Section~\ref{ObjectDetection} and describe the methods we've implemented for each task in Section~\ref{PPMethods}.

\section{Object Detection} \label{ObjectDetection}

The object detection model employed in this report primarily provides information regarding the sequence of the shuttlecock and the location of the players. Given that location in a badminton game is relative to the court and the net, the model also needs to detect these two elements. Additionally, the racket is detected to ensure the measurement of the habit foot. The five types of objects we aim to detect using the model are: \textbf{shuttlecock}, \textbf{player}, \textbf{court}, \textbf{net}, and \textbf{racket}.

We have chosen YOLOv7\cite{Wang_2023_CVPR} as our object detection model due to its impressive performance on the COCO dataset\cite{COCO}. Based on YOLOv7, the player, court, and net can be predicted with precision. While the racket is challenging to detect, the model still provides sufficient results to allow us to analyze which foot is the habitual one. However, detecting the shuttlecock is problematic due to its fast movement, which results in a blur effect, and its relatively small size in the frame.

Fortunately, TrackNet\cite{TrackNet} was introduced to detect objects such as the shuttlecock and tennis ball, both of which move rapidly and are small on-screen. During the training phase, we discovered that to detect these objects more precisely, a larger input size might be beneficial. Yet, this increase in size subsequently leads to longer training durations and higher computational demands. Most notably, when we consider using the maximum input resolution, equivalent to the original video dimensions of $1280\times 720$, the model becomes untrainable, even with a batch size set to $1$. 

\begin{figure}
\centering %图片居中
\includegraphics[width=0.45\textwidth]{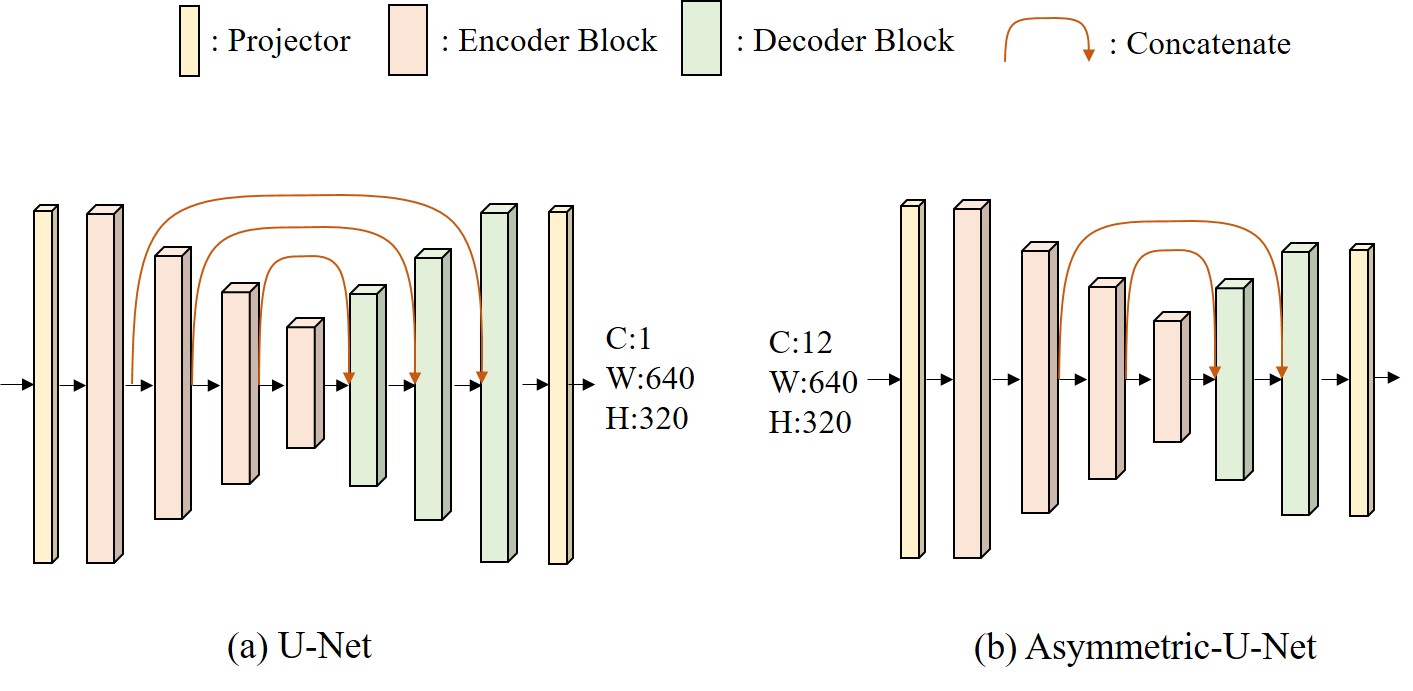}
\caption{U-Net and Asymetric U-Net} %最终文档中希望显示的图片标题
\label{fig.1} %用于文内引用的标签
\end{figure}

To improve the detection results for the shuttlecock, we decided to modify the original U-Net structure used in TrackNet to an Asymmetric U-Net\cite{TinySeeker}, as illustrated in Fig.\ref{fig.1}. Furthermore, to allow the model to learn features at different resolutions, we use a ground-truth heatmap to supervise outputs from every decoder block. The F1-score comparison is provided in Tab.\ref{tab.1}, and the computation cost comparison can be found in Tab.\ref{tab.2}. Experiments were conducted on a dataset we collected, comprising $128$ training videos and $17$ testing videos. Each video has frame-by-frame labeling.

\begin{table}[htbp]
\caption{The Comparison of F1-score on different model with input resolution $640\times 640$}
\begin{center}
\setlength{\tabcolsep}{1.8mm}{
\begin{tabular}{|c|c|}
\hline
Method & F1-score\\
\hline
U-Net & 0.93866 \\
\hline
Asym U-Net & 0.94255 \\
\hline

\end{tabular}
\label{tab.1}
}

\end{center}
\end{table}

\begin{table}[htbp]
\caption{The Comparison of FLOPs on different model with input resolution $640\times 640$}
\begin{center}
\setlength{\tabcolsep}{1.8mm}{
\begin{tabular}{|c|c|}
\hline
Method & FLOPs(G)\\
\hline
U-Net & 255.8 \\
\hline
Asym U-Net & 188.26 \\

\hline

\end{tabular}
\label{tab.2}
}

\end{center}
\end{table}

After predicting the shuttlecock's location throughout the entire video, we implemented a denoising process. This process involves removing ``jumping'' points using a median filter. Additionally, we filtered out points from the game's start and end using a straightforward timeline. We can reasonably assume the shuttlecock remains static during these periods, given that each video captures only one game. After the removal process, we applied interpolation to reconstruct the shuttlecock's trajectory. 

Up to this point, we have a set of shuttlecock locations. However, these results are not yet refined enough to directly make accurate predictions regarding ball type, hit number, and so on. Therefore, we need an additional post-processing model.

Let me add a supplementary note on the habit foot: we need to use VitPose\cite{VitPose} to predict human pose and combine it with the result of racket detection to determine habit foot.

\section{Post-process Methods} \label{PPMethods}

\subsection{Number of shots, Hit frame and Hitter}\label{Sec3.1}
The most straightforward method to detect the hitting time involves counting the inverse numbers of the moving direction on the y-axis. However, noise from the shuttlecock detections can easily lead to errors in these numbers. Consequently, we opted for an alternative approach: predicting whether ``Player A hits'', ``Player B hits'', or  from a given clip. This clip is generated using a sliding window of size $4$ that moves across the badminton video. For this task, we employed the X3D-M\cite{X3D} model, pretrained on the kinetics-400\cite{K400} dataset, to encode the spatiotemporal features necessary for these predictions. 

The labels for training the video analysis model are derived directly from the ground truth. For instance, if the 5th frame is marked in the ground truth as the frame where Player A hits, then the clips spanning frames $2 \to 6$, $3 \to 7$, and $4 \to 8$ are automatically labeled as ``Player A hits''.

After the X3D model produces its results, a sequence of predictions emerges. We then apply a smoothing filter to this sequence and select the most frequent category as the final prediction, yielding a sequence like [None, None, A, A, A, None, None]. In this context, None represents ``nothing happens'', A stands for ``Player A hits'', and B indicates ``Player B hits''. From the example sequence, the instance of ``Player A hits'' clearly occurs at the third frame.

\begin{figure}
\centering %图片居中
\includegraphics[width=0.43\textwidth]{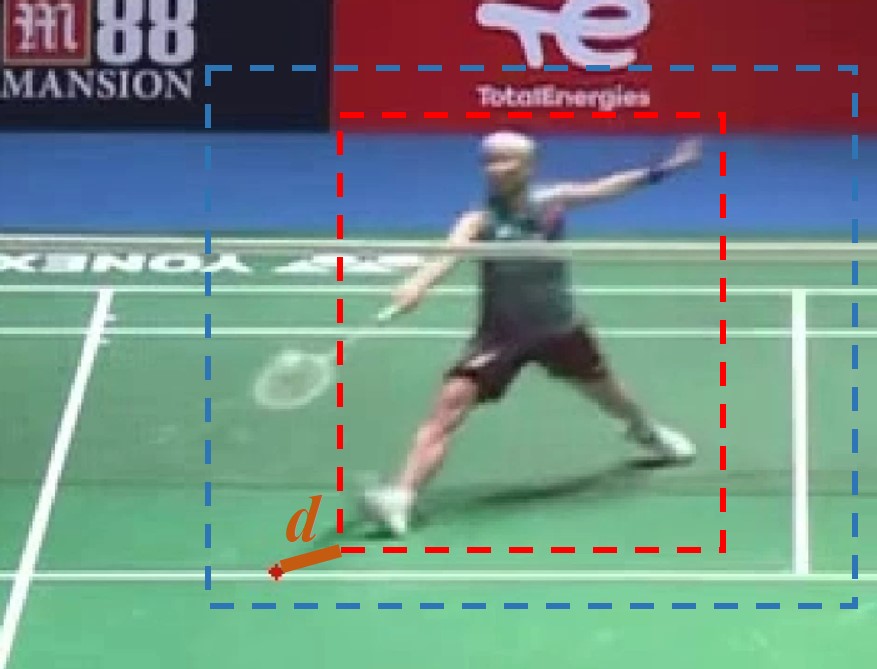}
\caption{The illustration of input data for the ``Location'' prediction.} %最终文档中希望显示的图片标题
\label{fig.2} %用于文内引用的标签
\end{figure}

Using this method, we can get results for the task \textbf{Number of shots}, \textbf{Hit frame}, and \textbf{Hitter}.

\subsection{BallHeight, LandingX/LandingY, HitterLocationX/HitterLocationY, DefenderLocationX/DefenderLocationY}

To predict ball height, the intuitive approach involves using the shuttlecock location at the hit frame and then mapping this location to the foot's y-axis. However, this method doesn't perform well due to noise in both the shuttlecock detection and human pose estimation. To address this, we employ EfficientNet\cite{EfficientNet} to predict the pixel offsets, denoted as dx and dy. These values, dx and dy, are determined by the ``CORNER'' and the ground truth. The term ``CORNER'' refers to the lower corner of the player's bounding box that is closest to the shuttlecock's location. 

The automatic labeling process is depicted in Fig.\ref{fig.2}. The red line represents the output from YOLOv7, while the blue line outlines a bounding box with dimensions $W\times 1.8$ and $H\times1.4$. We feed the region enclosed by the blue box into EfficientNet to predict the offsets dx and dy. We use this blue area because it encompasses both the shuttlecock and the projection point of the ground truth.

The method can be implemented for tasks \textbf{BallHeight}, \textbf{LandingX/LandingY},
\textbf{HitterLocationX/HitterLocationY}, and
\textbf{DefenderLocationX/DefenderLocationY}.

It's worth noting that we have tried various methods to determine dx and dy, such as the difference between the ground truth projection point and the shuttlecock, and the difference between the ground truth projection point and the player's center, etc. However, the most accurate and consistent results were achieved using the method mentioned earlier in this section.

\subsection{Backhand}
We directly use X3D to predict whether the player's action is a backhand or not. The training process is similar to that described in Section~\ref{Sec3.1}. We chose not to rely on human pose estimations for making predictions due to noise issues; VitPose cannot yield accurate results when the player is too blurry in the video.

\subsection{RoundHead}

\begin{figure}
\centering %图片居中
\includegraphics[width=0.45\textwidth]{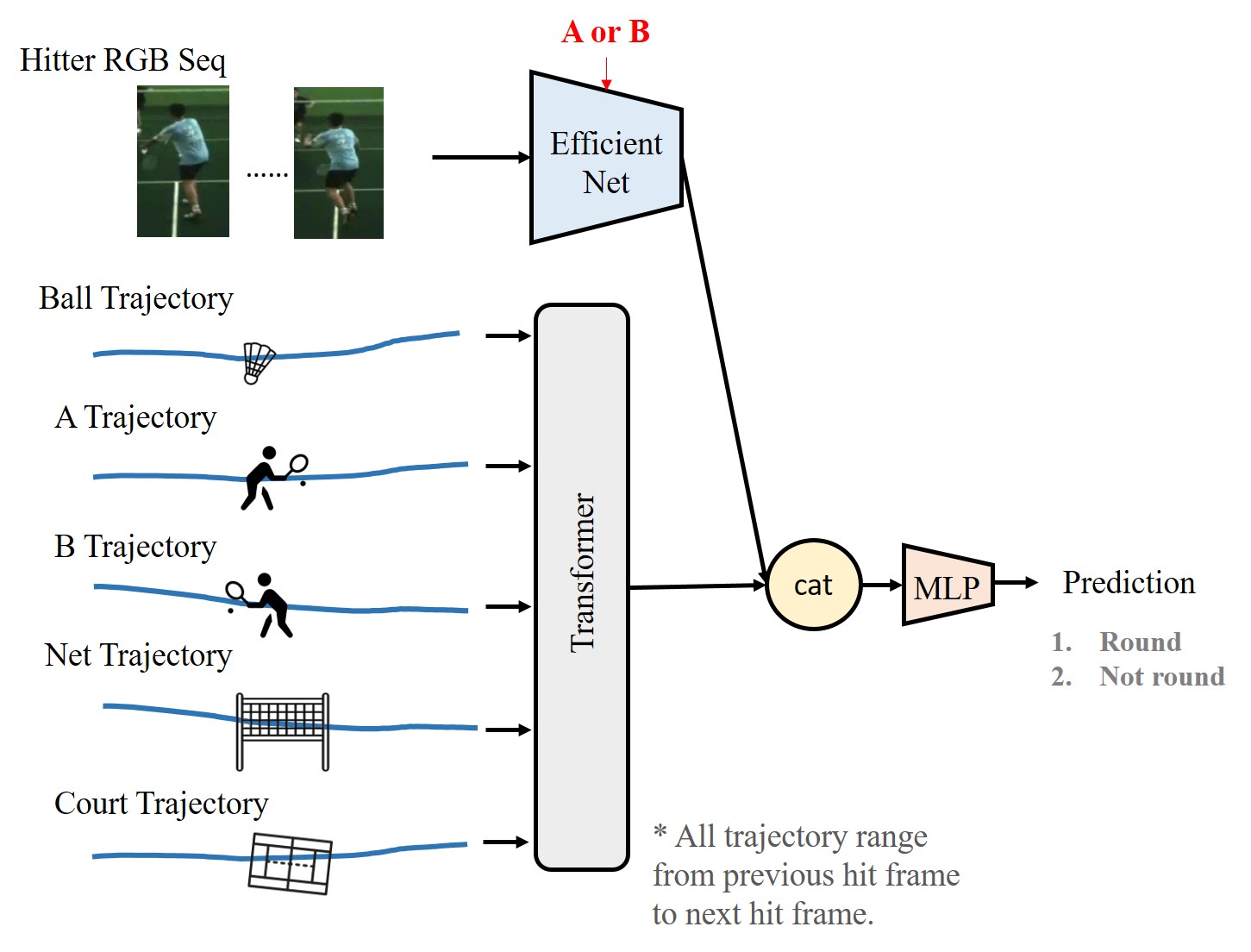}
\caption{The model of RoundHead task} %最终文档中希望显示的图片标题
\label{fig.3} %用于文内引用的标签
\end{figure}

The task of identifying a ``RoundHead'' action is more challenging than ``Backhand''. This complexity arises because the action can vary significantly, and its appearance may differ depending on the player's position relative to the camera. To mitigate this, we integrate sequences of the shuttlecock's location, the player's location, the court's location, and the net's location. As illustrated in Fig.\ref{fig.3}, we feed these sequences into a transformer\cite{Transformer} to extract location features. The insights derived from these location sequences are valuable; for instance, the relative positions of the player and the shuttlecock can influence the player's hitting pose. Notably, we employ a hint vector: ``Player A'' corresponds to [1,0] and ``Player B'' to [0,1]. This vector influences the X3D's output, as highlighted in Fig.\ref{fig.3} by the red text. We use a single fully connected layer to project this hint vector to match the feature dimensions.

\subsection{BallType}
We integrate sequences of the player's location, the court's location, and the net's location for prediction. We opted not to use the shuttlecock location sequence due to its inherent noise. For this task, the transformer doesn't use a mask to adapt to different input lengths. Instead, we set absent input tokens to $-100$, ensuring that fast shots and long shots have distinct input value ranges (fast shots should be smaller than long shots).

\begin{figure}
\centering %图片居中
\includegraphics[width=0.4\textwidth]{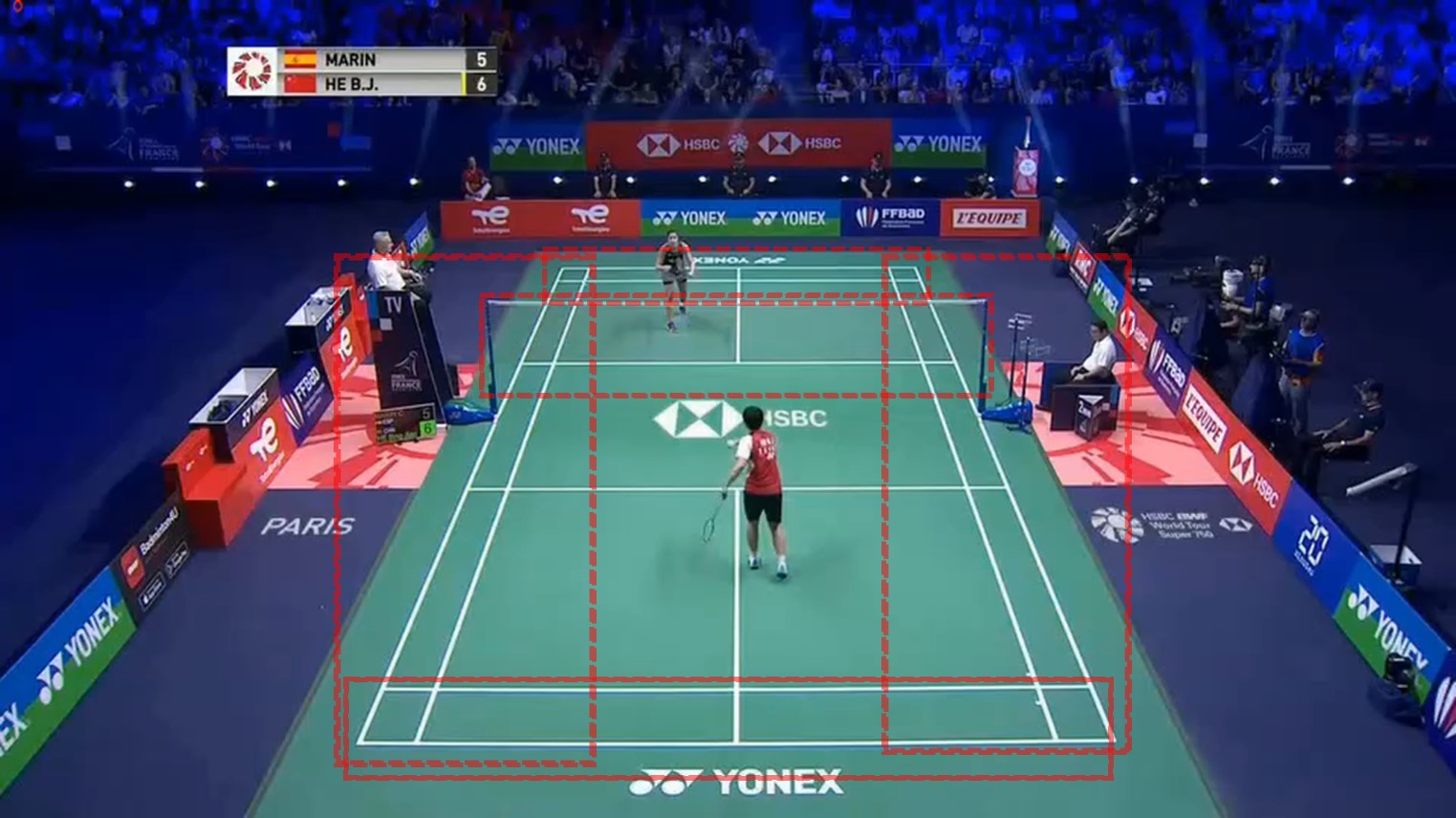}
\caption{The RoI for the input of the winner model} %最终文档中希望显示的图片标题
\label{fig.4} %用于文内引用的标签
\end{figure}

\subsection{Winner}
Directly using the court boundary and shuttlecock location can yield inaccurate results. We've previously discussed the inaccuracy of the shuttlecock's location. The potential inaccuracies in the court boundaries can be attributed to lighting conditions and camera angles. Therefore, we utilize the sequence from the last several RGB frames to determine the winner. To ensure the model remains focused and doesn't get sidetracked by extraneous details from the entire frame, we select the net area and the court's four boundary areas, as depicted in Fig.\ref{fig.4}, for input. This is analogous to having five cameras monitoring the game.

\subsection{Demoenstraction Video}
We have prepared some demonstration videos, available at the following URL: 

\url{https://www.youtube.com/shorts/BsOyQM44f28}

\section{Conclusion}

Benefiting from the wealth of annotated data provided by CoachAI Badminton 2023 Track1, we can employ a ``deeper'' learning strategy to address the inherent challenges. Our experiments demonstrated that, to the human eye, object detection results may seem satisfactory at first glance, but they don't always achieve the level of accuracy necessary for tasks focused on exact location determination. This highlights the crucial role of advanced deep learning algorithmic techniques in narrowing the disparity between human observational aptitudes and the exacting standards of our analytical tasks. As we look to the future, AI coaching systems hold the promise of numerous groundbreaking applications, poised to make a significant impact on our daily lives.

\bibliographystyle{named}
\bibliography{ijcai23}

\end{document}